# A Nonlinear Model for Time Synchronization

Frank Wang *and* Danjue Li, Equnix.com

## *Abstract*

The current algorithms are based on linear model, for example, Precision Time Protocol (PTP) which requires frequent synchronization in order to handle the effects of clock frequency drift. This paper introduces a nonlinear approach to clock time synchronize. This approach can accurately model the frequency shift. Therefore, the required time interval to synchronize clocks can be longer. Meanwhile, it also offers better performance and relaxes the synchronization process.

The idea of the nonlinear algorithm and some numerical examples will be presented in this paper in detail.

## I. INTRODUCTION

THIS document introduces a nonlinear approach for clock time synchronization, which includes adjustments to the offset and frequency. While the frequency correction takes longer time, and it is the major component of time synchronization.

Time synchronization can be treated as a time series event in machine learning. For each time step, the procedure includes: collection of timestamp data, learning of the clock system and correction of the clock time. The primary focus of this paper is to learn the clock using the timestamp data.

Existing approaches based on a linear model requires a smaller time interval to synchronize the clocks in order to achieve good performance. Here, we propose a nonlinear method which accurately models the frequency shift and therefore offers precise timing with fewer synchronizations. Furthermore, it can largely relax the synchronization process. A large time interval between adjacent synchronization means less information exchanges between clocks. It also allows for a large group of clocks to be synchronized simultaneously.

The main challenge of software-based timing is the noise and delay caused by various sources, which usually produces very noisy timestamp data. It is crucial to learn the clock system from the noisy data.

The nonlinear model works over much a longer time interval. Therefore, the longer time interval together with potential larger dataset make the model immune to noise and can largely reduce the variance of clock offset.

## II. LINEAR APPROXIMATION FOR NONLINEAR PROBLEM

In many processes, nonlinearities are not prominent, so their behavior can be described by simple linear models. In contrast, nonlinear models should be selected when strong nonlinearities are present. In general, a majority of dynamics are nonlinear. A nonlinear model captures the nature of the dynamics better than a linear model. Therefore, it performs better even with large steps in space and time.

For example, a straight line with non-zero slope can be approximately represented by a stepwise shape as shown in Fig. 1. The stepwise shape gets closer to the straight line when the number of steps increase. This illustrates the classical numerical approach, which splits large regions and domains into small regions and segments with the assumption of a constant value at each small region and segment. By analogy, we can think of the horizontal axis and vertical axis as time and clock frequency in time synchronization. The red and blue lines are similar as in the linear and nonlinear models in this paper.

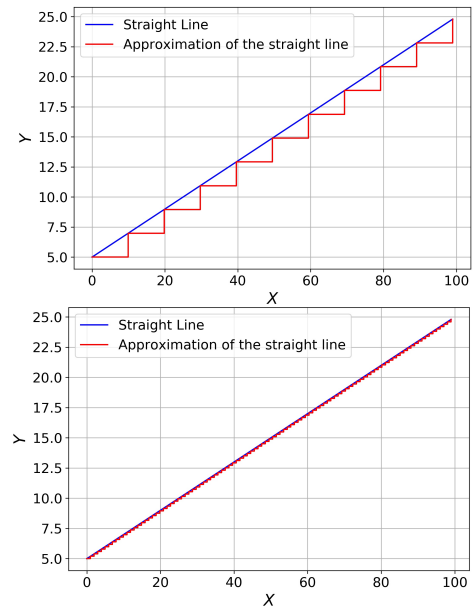

Fig. 1: Description of a straight line using the step-wise shape with 10 (top) and 100 steps (bottom).

When we say a clock is accurate, it actually means its frequency is accurate. The value of its frequency is precise and constant. In general, the clock frequency changes with time. Therefore, the clock time needs be adjusted more or less frequently based on the speed of frequency shift.

In general, clock time can be written as

$$T_c(t) = t_0 + \int_0^t (1 + s(t))dt + \theta. \qquad (1)$$

Where $t_0$, $\theta$, and $s$ are the starting time, offset, and skewness respectively. The skewness represents the deviate of the frequency and it changes with time.

In the real-world case, the integral in Eq. (1) is done numerically by choosing a short time step. The integral then becomes a summation over steps. The skewness is approximated by a fixed model at each step.

Existing time synchronizations use the linear model, which is based on the assumptions of a constant skewness $s(t) = \beta$ and offset during each time interval of clock synchronization. The clock time is giving by

$$T_c(t) = (1 + \beta)t + (\theta + t_0) \qquad (2)$$

The performance of linear model depends on the time step. Since the linear model assumes a constant clock frequency within each time step of clock synchronization, a small time-interval is required to synchronize the clocks in order to get better performance.

Although the linear model is easy to implement and works reasonably well with a small enough time step, it requires frequent time synchronization. During each synchronization event, enough communication between clocks is needed to remove the noise and jitter. However, the frequent communication increases the network load. Moreover, the communication data becomes limited for short time intervals, making the performance of synchronization algorithm limited and largely impacted by the noise and errors from the measurement. As a result, the frequent synchronization usually has large jitter in the clock synchronization, which is shown by a large standard deviation in the clock offset. While the nonlinear model works with a long time window, the effect of noise and jitter can be largely reduced by smoothing and averaging.

## III. DYNAMICS OF CLOCK FREQUENCY CHANGE

To synchronize the clocks, it is critical to understand the dynamics of clocks and the impact from their environments. How frequently the clocks should be synchronized depends largely on the pattern of clock frequency drift and the algorithm chosen. The property of clock frequency is briefly discussed here.

The frequency of an oscillator slowly changes over time [3]. The frequency shifts about 1.0 ppm per day in the example shown in Fig. 2. The shift pattern changes approximately linearly with time in one day period.

The frequency of the clock also varies with temperature. Crystal frequency characteristics depend on the shape or cut of the crystal. A tuning-fork crystal is usually cut such that its frequency dependence on temperature is quadratic with the maximum around 25 °C [4]. This means that a tuning-fork crystal oscillator resonates close to its target frequency at room temperature, but the frequency drops slowly when the temperature either increases or decreases from room temperature. A common parabolic coefficient for a 32 kHz tuning-fork crystal is −0.04 ppm/°C².

$$f = f_0[1 - 0.04ppm/°C^2(T - T_0)^2] \qquad (3)$$

One measurement [5] is shown in Fig. 3.

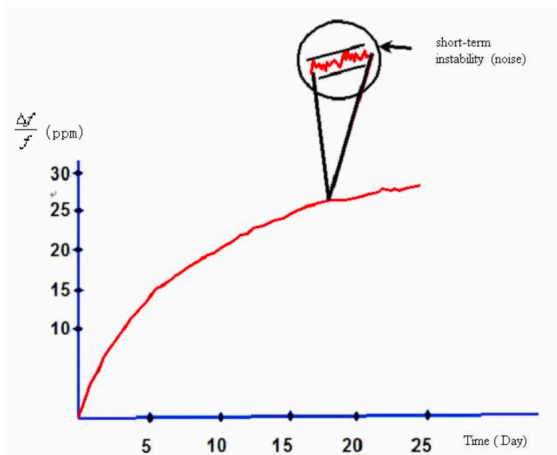

Figure 2: Aging of crystal resonator reference [3].

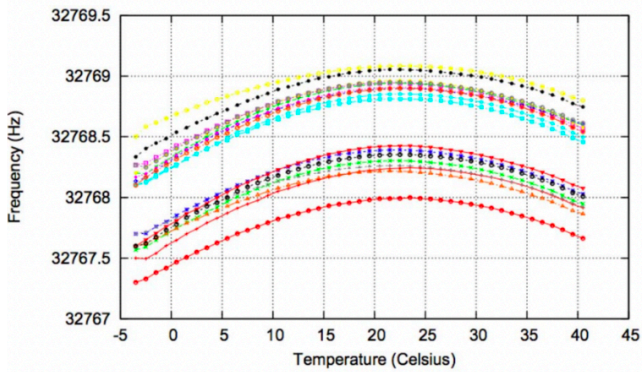

Fig. 3. Measured temperature effect on the frequency of several crystal oscillators (nominal frequency f0=32768.5Hz), reference [5]

However, linear approximation is good model for a small variation of 5 Degrees. Note that the room temperature in a data center and most office building is well controlled, and it usually varies slowly. Therefore, it is safe to use a linear approximation to model the temperature effect within a time window of a few hours.

In summary, the frequency of a clock is its intrinsic property, and the environment has little impact on it except in extreme cases (Otherwise, it would not be proper to be used as a clock!). For instance, the frequency of a pendulum clock depends on the length of the pendulum and the acceleration of gravity. A linear model is a good approximation to represent the frequency shift due to both the aging and temperature effects within a time window of a few hours.

## IV. NONLINEAR MODEL IN TIMING

The linear model works well when the time interval between the adjacent synchronization process is small enough, so a nonlinear process can be treated as a linear process, which means that the clocks are required to be synchronized more frequently. This potentially takes large bandwidth for communication among clocks.

The nonlinear model includes the variation of skewness with time. It assumes that the skewness changes with time linearly or in higher order form. For example, if the skewness changes linearly with time, $s(t) = \alpha t + \beta$ , the clock time can be simplified as

$$T_c(t) = \frac{1}{2}\alpha t^2 + (1+\beta)t + (\theta + t_0). \quad (4)$$

The first term represents the frequency shift effect. In a real-world case, the skewness changes approximately in a linear way within an intermediate time interval.

On the other hand, the nonlinear model actually outperformances the linear model even with a large time step because it accurately captures the frequency shift with time. When the time step is small enough, the linear model gets a similar performance as the nonlinear model.

When clock frequency drifts slowly, the linear model is a good choice because of its easy implementation. However, a small time step is required for a clock system with fast frequency drift. This can become unworkable if the clock system doesn't have enough time to communicate and synchronize with the clocks. In this case, a nonlinear model is the only choice to get high performance synchronization.

The large time step of the nonlinear model also offers a number of other advantages:

- Smooth and continuous time correction.
- Collection of large clock data sets. This can potentially improve the performance of the algorithm by increasing the accuracy of the model in statics and filtering bad data.
- Reduction of network traffic induced by the communication with server clocks.
- Allows for a sophisticated algorithm to optimize a large clock graph; for example, a system with hundreds or even thousands of clocks.

Fig. 4 shows an example of PTP4L results where the constant frequency correction is removed to clearly compare it with the clock's offset. It is important to note that the frequency correction limits the resolution of the clock offset. It shows a strong correlation between the clock offset and the frequency correction. The calculated correlation factor is 0.96. In other words, the fluctuation in the offset is mostly derived from the frequency correction $\delta_{offset} \approx \delta_{fre}\Delta t$. The variance of offset is about proportional to the variance of the frequency. The nonlinear model corrects frequency continuously and smoothly and therefore there is a smooth correction in the offset too. Since the nonlinear model learns the frequency in a better way, it offers continuous, smooth, and accurate timing correction. For Fig 4, the clock offset by a nonlinear model will be smooth and closer to zero as illustrated by the green line. The variance of time offset drops significantly. This is a distinct advantage over the linear model.

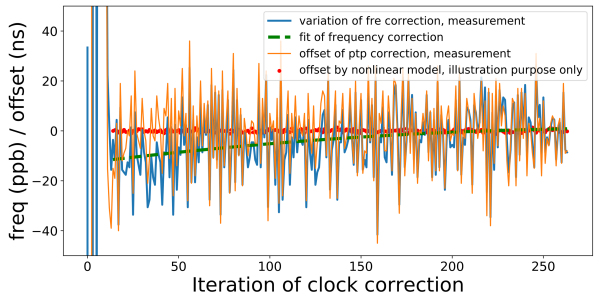

Fig.4. Measured frequency correction (blue) and clock offset (yellow) in a ptp4l test. The constant part of the frequency correction is removed from the data for better comparison. The red dots and green line are for illustration purposes only, which is not real data.

## V.    APPLICATION TO PTP CASE

### A.  Timestamp data

The nonlinear model can be used in various cases whenever the frequency of the clock is used. Here we use PTP data for demonstration purposes only. The core of the protocol is summarized in the following picture:

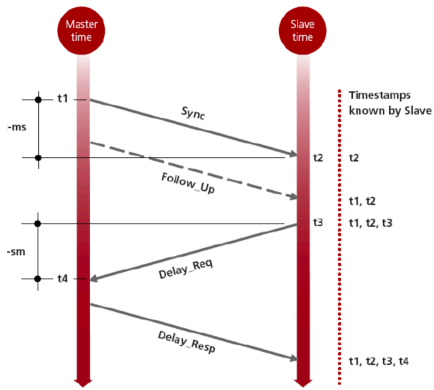

Fig. 5. Signaling between two clocks. t1 and t2 are the corresponding transmit and receive timestamps. t3 and t4 are for the reverse direction.

With two steps communication between the two clocks, the nonlinear model, the difference in timing between the clocks forms bands ($t3 - t4, t2 - t1$). Fig. 6 shows one numerical example. The data and plot are the same as the linear model, except with a much longer time interval. Therefore, the nonlinearity becomes important.

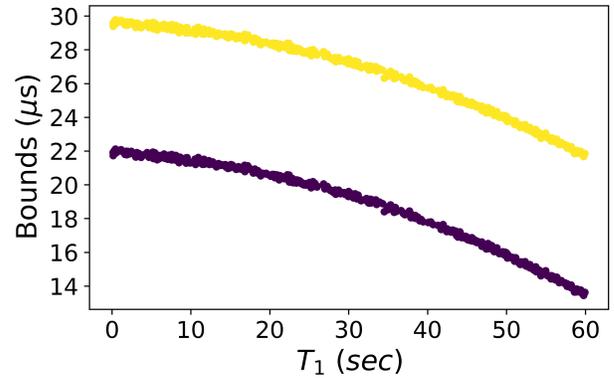

Fig. 6.  Bounds on discrepancy between clocks

### B.  Learn clock parameters

The first and second order derivatives of clock discrepancy can be learned using various algorithms; for example, high order SVM (Support Vector Machine) [1] as shown in Fig.7. Colors represent the regimes (bounds) learned by the algorithm. The slope and offset give the skewness and offset of the clock, respectively. The first and second order derivatives can be derived from the separation plane. For the nonlinear model, there is a longer time interval to synchronize the clocks and therefore, if necessary, we can collect a larger number of data points to improve the accuracy of the model.

The linear SVM is immune to the noise and jitters far from the separation plane as shown in Fig. 8. However, the nonlinear model performs poorly with the large noise and jitter. Instead of working with outlier, regression with inliers is helpful, but it increases the complexity of learning.

Several data processing approaches can be applied to improve the performance of the learning: such as packet selection, data filtering and other advanced techniques.

There is better and easier way to handle such problem in our case. It's important to note that the difference between the two bounds gives the propagation distance $d$ (the distance between the two clocks) as:

$$d = \frac{(t2 - t1) - (t3 - t4)}{2} \qquad (5)$$

It is obvious that all message should take a time equal to (without additional delay) or longer than true physical distance to reach another clock. Therefore, a data point of the upper bound (red dots in Fig. 8) is bad data if it is far below the upper bound support line. Similarly, a data point of the lower bound (yellow dots in Fig. 8) belong to bad data if it is far above the lower bound. After such data clean, the outliers in such clean data have no impact on the learning in SVM. Therefore, the learning can be improved by applying this constrain.

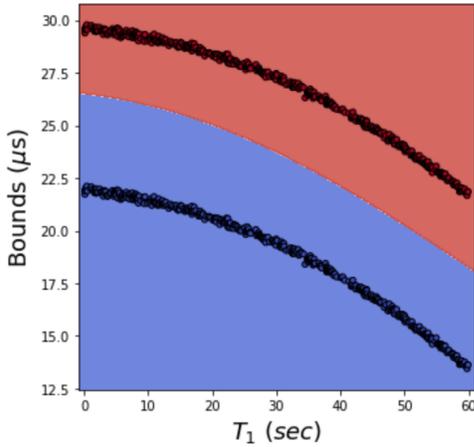

Fig. 7. Nonlinear prediction on the timestamp data

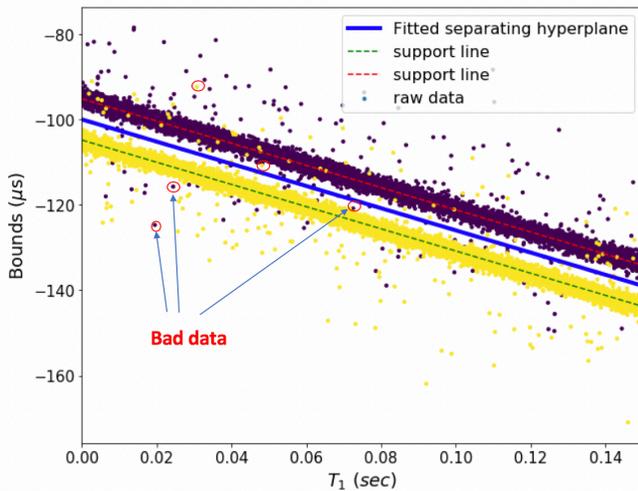

Fig. 8. Working with noisy timestamp data using Linear SVM model. A few bad data points are marked for illustration.

## C. Numerical convergence test

In this example, two clocks are synchronized based on the two-way exchange scheme shown in Fig. 5. During each time interval for synchronization (called time step in this paper), the two clocks communicate 2000 times, no matter what time step is used. Noise is added to represent the jitter in the delay. The noise has the normal distribution with an *rms* of 0.01 $\mu s$. The nonlinear model learns up to *2nd* order (Eq. 1). The model learned in previous step is used to synchronize the clock. Fig. 9 shows the synchronization with various time steps, from 2 *seconds* to 100 *seconds*. Note that there is no synchronization for the first 3 steps. Instead, the algorithm uses that period to learn the system. The nonlinear model works well, and it always converges quickly in a few steps. For a large time step of 100 *seconds*, there is over-correction in the first 2-3 steps since we don't optimize the algorithm. But the algorithm still converges well.

It is a concern that it takes a long time to converge for a long time-step because it typically takes 10 to 20 times steps to converge. A hybrid time-step model is proposed. A small time-step is chosen at the beginning for fast convergence and then a long time-step is adapted. Fig. 10 shows two examples with the hybrid time-step. A time-step of 2 seconds is used at the beginning of the synchronization. The hybrid model always converges within 20 seconds. Afterwards, a long time-step of 200 seconds works smoothly. This hybrid model also eliminates the spikes (over-correction) for long time-step cases as in Fig. 10.

For comparison, the linear model is shown in Fig. 11. A time-step of 2 sec works well, but not for 10 sec where there is a large error. Note that the linear model converges slowly but smoothly. In all the tests, the same number of communications between the clocks at each step are used. For the nonlinear model with a large time step, there is an option to improve the performance by increasing the number of commutations between clocks if the performance is degraded by noise and jitter.

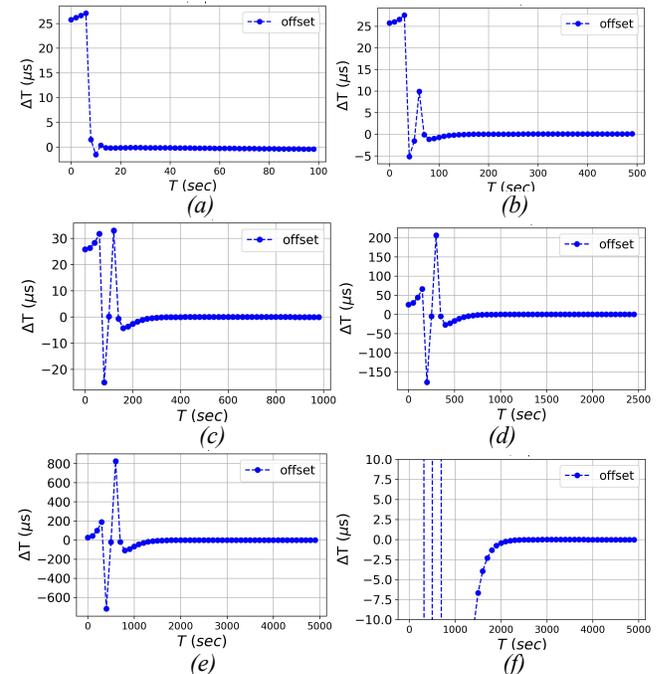

Fig. 9. Numerical test on nonlinear synchronization with various time steps in synchronization (a) *2 seconds*, (b) *10 seconds*, (c) *20 seconds*, (d) *50*

*seconds*, (e) and (f) *100 seconds* with different scale. The vertical axis shows the offset.

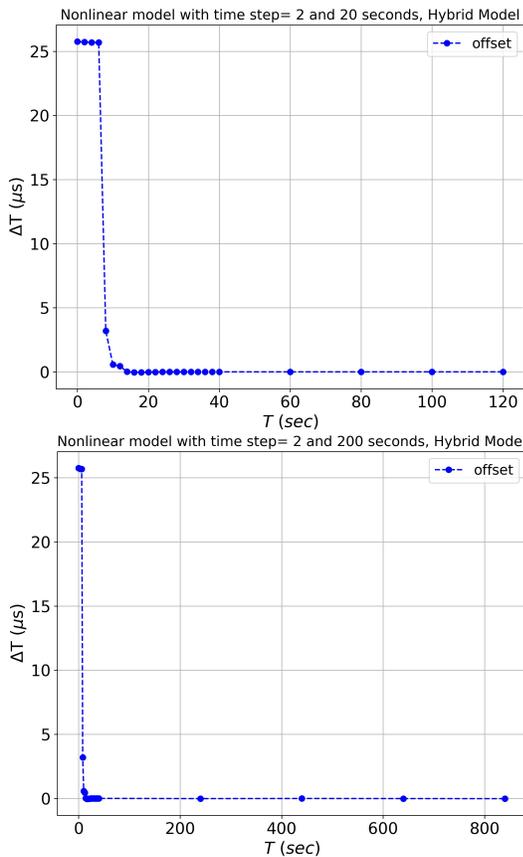

Fig. 10. Numerical test on nonlinear synchronization with hybrid time steps (top) *2 and 20 seconds*, (bottom) 2 and *200 seconds.*

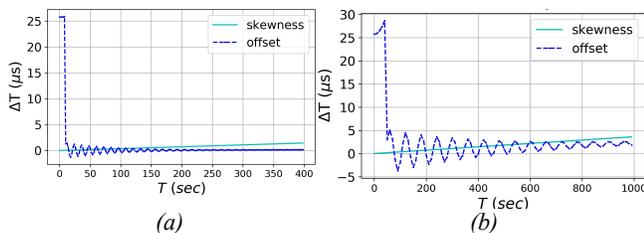

*(a)*                                    *(b)*

Fig. 11. Numerical test on linear synchronization with time step of 2 *seconds* (a) and 10 *seconds* (b).

### D. Effect of noise and jitter from measurement

Precise timing requires both good timing sources (for instance GPS) and accurate measurement. The hardware solution provides precise timing but it's expensive and not scalable. The software method provides a cheap and scalable solution. However, it suffers from noise and jitter from the measurement, which fundamentally limits the performance of the software method.

For instance, the measured frequency oscillates at a high frequency as shown in Fig. 4. In reality a clock's frequency should be stable in such a short period. The linear model works only for a very short time-step, while the nonlinear method we proposed works in longer period time window. The noise and jitter are smoothed out when the nonlinear model works with large dataset in a longer period of time. The frequency in nonlinear model is close to the green line in Fig. 4. In general, the nonlinear model is immune to the noise and jitters. The model is free of the noise with a frequency larger than $1/\Delta t$. The larger the time step, the more robust the model. If the noise is purely random, it has no impact on the nonlinear model.

### E. Queueing effect due to network load

One of the important jitter sources is the queueing effect due to network traffic. When the network load is high, the communication between the servers are delayed in a complicated way. The higher the network load, the larger the delay. The direct ways to eliminate the impact of queueing effect includes a dedicated line for timing or setting high priority authentication for timing communication.

Nevertheless, we briefly discuss the queueing effect here. The in-bound and out-bound network traffic can be different. Therefore, the different delay by queueing induces extra asymmetry between the paths of in-bound and the out-bound traffic. If the algorithm for synchronization can't handle this asymmetry, timing errors will occur. Most existing algorithms assume symmetry for the in-bound and out-bound paths. Therefore, the queueing effect is an issue for high network load cases. An algorithm which is capable of handling dynamic asymmetry is desired for both the linear and nonlinear methods. We are investigating potential way to handle the queuing effect.

Aside from the path asymmetry, the network load also increases the noise level in the measurement. Similar as other type noise, the nonlinear model is better in handling noise, especially for a steady noise pattern.

However, the network load changes with time. Since the nonlinear model works over a longer time window, it will average out the queueing effect in time if the queueing induces fast variation in path length. On the other hand, the linear model generates oscillation in timing which correlates with the pattern of queueing effect in time.

## VI. Summary

A nonlinear method was proposed and tested numerically for time synchronization. It handles the clock frequency shift in a better way and provides precise timing with large time interval for synchronization. The linear model works only with small time steps (order of seconds, the more frequent the synchronization). Therefore, the traffic load induced by the message exchange and the computation time is largely reduced. Meanwhile, the nonlinear algorithm also converges faster and is immune to the noise and jitter from the measurement. This is due to the large time window and the data. This largely reduces the variance of clock offset.

## VII. References

[1] Corinna Cortes and Vladimir Vapnik (1995). "Support-vector networks". Machine Learning, 20 (3):273–297. doi:10.1007/BF00994018.

[2] Yilong Geng, Shiyu Liu, Zi Yin, Ashish Naik, Balaji Prabhakar, Mendel Rosenblum, Amin Vahdat, Exploiting a Natural Network Effect for Scalable, Fine-grained Clock Synchronization, 15th USENIX Symposium on Networked Systems Design and Implementation, NSDI'18

[3] John R. Vig, "Quartz Crystal Resonators and Oscillators for Frequency Control and Timing Applications - A Tutorial", 2004 IEEE International Frequency Control Symposium Tutorials, May 2004

[4]https://en.wikipedia.org/wiki/Crystal_oscillator#Temperature_effects

[5] Martin Lévesque and David Tipper, "A Survey of Clock Synchronization Over Packet-Switched Networks", IEEE COMMUNICATIONS SURVEYS & TUTORIALS, VOL. 18, NO. 4, FOURTH QUARTER 2016